\documentclass[letterpaper]{article} 
\usepackage{aaai25}  
\usepackage{times}  
\usepackage{helvet}  
\usepackage{courier}  
\usepackage[hyphens]{url}  
\usepackage{graphicx} 
\usepackage{natbib}  
\usepackage{caption} 
\usepackage{algorithm}
\usepackage{algorithmic}

\usepackage{multicol, multirow, threeparttable}
\usepackage{amsmath}
\usepackage{amsthm}
\usepackage{amssymb}
\usepackage{booktabs}
\usepackage{pifont}
\usepackage{newfloat}
\usepackage{listings}
\DeclareCaptionStyle{ruled}{labelfont=normalfont,labelsep=colon,strut=off} 
\floatstyle{ruled}
\newfloat{listing}{tb}{lst}{}
\floatname{listing}{Listing}
\usepackage{subcaption}
\usepackage{xspace}
\usepackage{xcolor}         
\usepackage{marvosym}

\definecolor{bl}{rgb}{0.25, 0.5, 0.9}
\newcommand{\best}[1]{{\textbf{\textcolor{red}{#1}}}}
\newcommand{\second}[1]{{\textcolor{bl}{\underline{#1}}}}

\newcommand{\circled}{\ding{172}}
\newcommand{\circledd}{\ding{173}}
\newcommand{\circleddd}{\ding{174}}

\newcommand{\NAME}{CALF\xspace}

\usepackage[capitalize]{cleveref}
\crefname{section}{Sec.}{Secs.}
\Crefname{section}{Section}{Sections}
\Crefname{table}{Table}{Tables}
\crefname{table}{Tab.}{Tabs.}

\title{CALF: Aligning LLMs for Time Series Forecasting via Cross-modal Fine-Tuning}
\author{
    Peiyuan Liu\textsuperscript{\rm 1,}\equalcontrib,
    Hang Guo\textsuperscript{\rm 1,}\equalcontrib,
    Tao Dai\textsuperscript{\rm 2,\thanks{Corresponding author: Tao Dai and Naiqi Li.}}, 
    Naiqi Li\textsuperscript{\rm 1,\footnotemark[2]}, 
    Jigang Bao\textsuperscript{\rm 1}, \\
    Xudong Ren\textsuperscript{\rm 1}, 
    Yong Jiang\textsuperscript{\rm 1}, 
    Shu-Tao Xia\textsuperscript{\rm 1,3}
}
\affiliations{
    \textsuperscript{\rm 1}Tsinghua Shenzhen International Graduate School \\
    \textsuperscript{\rm 2}Shenzhen University \\
    \textsuperscript{\rm 3}Pengcheng Laboratory
 \\
    \{lpy23, guo-h23\}@mails.tsinghua.edu.cn, \{daitao.edu, linaiqi.thu\}@gmail.com
    
}

\usepackage{bibentry}

\begin{document}

\maketitle

\begin{abstract}
Deep learning (e.g., Transformer) has been widely and successfully used in multivariate time series forecasting (MTSF). Unlike existing methods that focus on training models from a single modal of time series input, large language models (LLMs) based MTSF methods with cross-modal text and time series input have recently shown great superiority, especially with limited temporal data. However, current LLM-based MTSF methods usually focus on adapting and fine-tuning LLMs, while neglecting the \textit{distribution discrepancy} between textual and temporal input tokens, thus leading to sub-optimal performance. To address this issue, we propose a novel \textbf{C}ross-Mod\textbf{A}l \textbf{L}LM \textbf{F}ine-Tuning (\NAME) framework for MTSF by reducing the distribution discrepancy between textual and temporal data, which mainly consists of the temporal target branch with temporal input and the textual source branch with aligned textual input. To reduce the distribution discrepancy, we develop the cross-modal match module to first align cross-modal input distributions. Additionally, to minimize the modality distribution gap in both feature and output spaces, feature regularization loss is developed to align the intermediate features between the two branches for better weight updates, while output consistency loss is introduced to allow the output representations of both branches to correspond effectively. Thanks to the modality alignment, \NAME establishes state-of-the-art performance for both long-term and short-term forecasting tasks with low computational complexity, and exhibits favorable few-shot and zero-shot abilities similar to that in LLMs.
\end{abstract}

\begin{links}
    \link{Code}{https://github.com/Hank0626/CALF}
\end{links}


%

\section{Introduction}

Multivariate time series forecasting (MTSF) plays a crucial role in the domain of time series analysis and has further boasted a wide range of real-world applications including weather forecasting~\citep{angryk2020weather}, energy prediction~\citep{demirel2012forecasting}, financial modeling~\citep{patton2013copula}. 
To achieve more accurate forecasting performance, numerous deep learning-based MTSF methods trained on a single modal of time series input have been developed in recent years~\citep{wu2023timesnet, zhang2022crossformer, nie2022pathtst, zeng2023dlinear, zhou2022fedformer, wen2022survey, dai2023periodicity, duet} and have gained great success. 

However, previous single-modal MTSF methods may suffer from overfitting problems, due to the limited training data, thus limiting their real applications. To relieve such issues, some pioneering works attempt to introduce the powerful Large Language Models (LLMs) models in time series forecasting by employing the strong context modeling ability of LLMs. For example, Zhou et al. \citep{zhou2023onefitsall} proposed a unified time series analysis framework by adapting and fine-tuning LLMs. Building upon this, other works have introduced additional enhancements to further expand the capabilities of LLMs in time series forecasting, including refining fine-tuning methods \citep{chang2023llm4ts}, sequence decomposition \citep{cao2023tempo}, and the incorporation of textual prompts \citep{jin2023timellm}. Benefiting from the large-scale pre-training, LLM-based methods not only exhibit strong context modeling capabilities but also help mitigate the problem of overfitting.

\begin{figure*}
    \centering
    \includegraphics[width=0.95\textwidth]{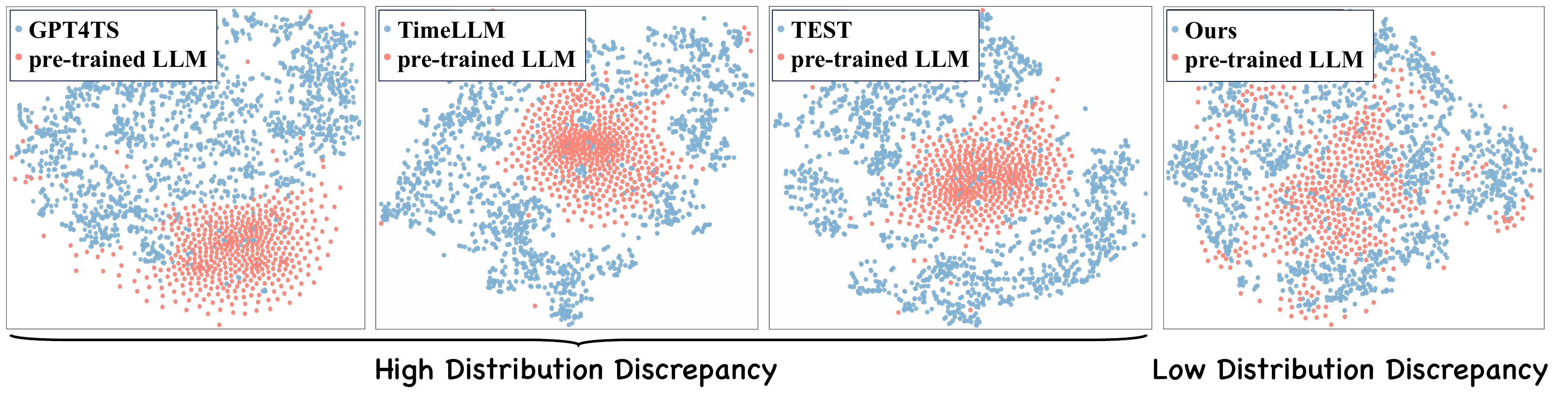}
    \caption{The t-SNE visualization of pre-trained word token embeddings of LLM with the hidden features from the penultimate layer from GPT4TS \cite{zhou2023onefitsall}, TimeLLM~\cite{jin2023timellm}, TEST~\cite{sun2024test}, and ours of ETTh2 dataset. Current LLM-based methods either use linear layers to project time series to the LLM's feature dimension \cite{zhou2023onefitsall} or employ cross-attention and contrastive learning techniques \cite{jin2023timellm, sun2024test}, which address only the input side and overlook alignment in the deeper layers. Our CALF achieves better alignment through multi-level cross-modal fine-tuning.}
    \label{fig:intro}
\end{figure*}

Despite the great success of LLM-based MTSF methods, existing methods suffer unfavorable results on common MTSF benchmarks, as well as pool performance on few-shot or zero-shot tasks. As shown in \cref{fig:intro}, we visualize the distribution of the textual and temporal tokens of existing LLM-based MTSF methods, and it can be seen that the temporal tokens in existing LLM-based methods cannot align well with the original textual tokens from LLMs \cite{zhou2023onefitsall, sun2024test, jin2023timellm, chang2023llm4ts}. As a result, this distribution discrepancy contributes in part to the sub-optimal performance of the current approaches. We argue that the main reason for the above misalignment is due to the fact that existing methods only investigate the input side to match the feature dimensions between time series and LLM, and this simple solution hinders the manipulation of LLM's powerful generalization capabilities. The above observations motivate us to develop a multi-level cross-modal aligning framework to alleviate the distribution discrepancy between temporal modal input and textural modal weights.


Inspired by the above observations, we propose a \textbf{C}ross-Mod\textbf{A}l \textbf{L}LM \textbf{F}ine-Tuning (\NAME) framework, which employs cross-modal fine-tuning to allow more comprehensive alignment between temporal target modalities and textual source modalities. Specifically, \NAME consists of two branches: the temporal target branch and the textual source branch. The temporal target branch processes time series information, while the textual source branch extracts and adapts information from pre-trained LLMs using aligned textual modal tokens. To bridge the modality gap between these branches, we introduce three meticulously designed cross-modal fine-tuning techniques (see \cref{fig:pipeline}): (1) \textbf{Cross-Modal Match Module} integrates time series and textual inputs through principal word embedding extraction and a cross-attention mechanism, ensuring efficient alignment of the marginal input distribution between time series and text; (2) \textbf{Feature Regularization Loss} aligns the outputs of each intermediate layer, ensuring that gradients at every layer are more effectively guided for better weight updates; (3) \textbf{Output Consistency Loss} ensures that the output representations of textual and temporal series modalities correspond effectively, resolving discrepancies in the representation space and maintaining consistent semantic context for time series data. Through a more comprehensive alignment, our \NAME consistently achieves state-of-the-art performance in both long-term and short-term forecasting across multiple datasets, demonstrating excellent few/zero-shot generalization capabilities, while maintaining significantly low complexity. 

The contributions of this paper are threefold:
\begin{enumerate}
    \item We identify the significant distribution discrepancies between textual and temporal modalities in existing LLM-based forecasting models and highlight the importance of addressing this misalignment for improved performance.
    \item We propose \NAME, a novel framework that employs cross-modal fine-tuning techniques to comprehensively align temporal and textual data. The framework includes three specific methods: the Cross-Modal Match Module for aligning input distributions, Feature Regularization Loss for better gradient guidance and weight updates, and Output Consistency Loss for resolving output representation space discrepancies and maintaining consistent semantic context.
    \item Extensive experiments on eight real-world datasets demonstrate that \NAME achieves state-of-the-art performance on both long-term and short-term time series forecasting tasks, with favorable generalization ability and low computational complexity.
\end{enumerate}

\section{Related Work}

\subsection{Time Series Forecasting}

In recent years, deep learning has significantly revolutionized the field of time series forecasting, with a plethora of methods emerging to enhance predictive accuracy~\citep{zeng2023dlinear, das2023tide, wu2023timesnet, wang2022micn, liu2023wftnet}. Among these, Transformer-based models have emerged as the frontrunners, offering unparalleled performance due to their exceptional ability to model complex dependencies in data ~\citep{nie2022pathtst, zhang2022crossformer, woo2022etsformer, wu2021autoformer, zhou2022fedformer, dai2023periodicity}. However, they often have limitations due to the scarcity of training data, overfitting in specific domains, and the necessity for intricate architectural designs.

To tackle these challenges, the integration of LLMs into time series forecasting has emerged as a novel and promising direction. This approach leverages the extensive pre-training of LLMs to enhance the context-modeling capacity in time series analysis. A groundbreaking framework proposed by Zhou et al. \citep{zhou2023onefitsall} first demonstrated the potential of adapting LLMs for time series analysis. Following this paradigm, subsequent research has introduced further refinements and innovations. For example, Chang et al. \citep{chang2023llm4ts} introduced a novel two-stage fine-tuning method and integrated time-series patching with additional temporal encoding into pre-trained LLMs. Cao et al. \citep{cao2023tempo} incorporated decomposition of time series and selection-based prompts for adapting to non-stationary data. However, these works often directly input time series data into LLMs, overlooking the misalignment between time series and textual modalities. Some works have attempted to address this issue. Sun et al. \citep{sun2024test} aligned time series data with LLM embeddings using contrastive learning and employed soft prompts for effective time series task handling. Jin et al. \citep{jin2023timellm} reprogrammed time series input with text prototypes and enriches it using context as a prefix for LLM alignment. Despite these efforts, the alignment strategies have not been sufficiently effective.

\subsection{Cross-Modal Fine-tuning}
The objective of cross-modal fine-tuning is to apply models pre-trained on data-rich modalities to data-scarce modalities, addressing issues of data insufficiency and poor generalization \cite{shen2023orca}. Many existing works focus on transferring LLMs to other modalities, such as vision \cite{kiela2019vision, verma2024vision1}, audio \cite{jin2023cross, hassid2024speak}, and biology \cite{vinod2023bio, xiao2021bio1}. These efforts provide initial evidence of the cross-modal transfer capacity of pre-trained models. In the domain of time series, current research primarily leverages the powerful contextual modeling capabilities of LLMs to fine-tune them for improved forecasting performance \cite{zhou2023onefitsall, jin2023timellm, cao2023tempo, chang2023llm4ts, sun2024test}, often neglecting the gap between the input and output distributions of language and time series modalities. In this work, we apply cross-modal fine-tuning techniques to address the challenge of transferring pre-trained language model knowledge to the time series modality.

\section{Methodology}

\begin{figure*}[!t]
\centering
\includegraphics[width=1\textwidth]{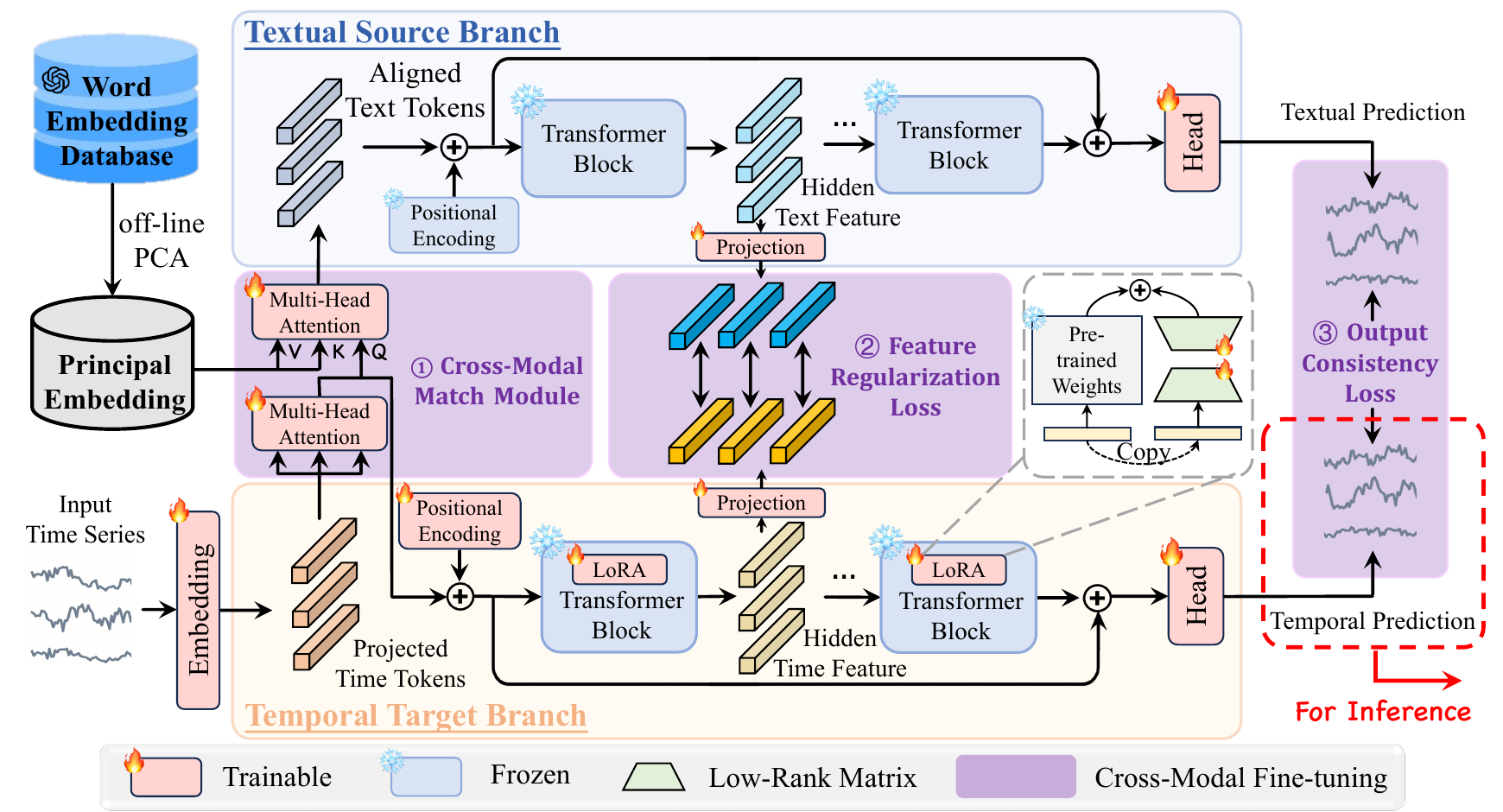}
\caption{An overview of the proposed cross-modal fine-tuning framework. Above is the Textual Source Branch, and below is the Temporal Target Branch. To bridge the modality gap, the framework employs three cross-modal fine-tuning techniques: \circled  Cross-Modal Match Module, \circledd Feature Regularization Loss, and \circleddd  
 Output Consistency Loss.}
\label{fig:pipeline}
\end{figure*}

As shown in~\cref{fig:pipeline}, our proposed \NAME consists of two branches: the textual source branch and the temporal target branch. In concrete, the textual source branch takes the aligned text tokens $X_{text}$ as input and employs $L$ stacked pre-trained LLM layers to obtain the hidden text feature $F^{l}_{text}$, where $l=\{1, \cdots , L\}$. A task-specific head is used to generate the output $Y_{text}$. Meanwhile, the temporal target branch works with the projected time series tokens $X_{time}$, and uses the same number of layers $L$ with identical pre-trained weights as the textual source branch to obtain the hidden time feature $F^l_{time}$. The output of this branch is denoted as $Y_{time}$. To bridge the modality gap between these two branches, we utilize three cross-modal fine-tuning techniques to fine-tune the temporal target branch: the \textbf{Cross-Modal Match Module}, the \textbf{Feature Regularization Loss}, and the \textbf{Output Consistency Loss}. Detailed descriptions of these techniques will be provided in the following section.


\subsection{Cross-Modal Match Module}

As demonstrated in previous work~\citep{mikolov2013efficient}, the matrices of word embedding layers in pre-trained LLMs constitute a well-structured context representation space, \textit{e.g.}, semantic distances between different words can be quantified through vector similarity. This word embedding layer represents the input distribution of the language modality in pre-trained LLMs. Despite this promising property, previous LLM-based time series methods often overlook this distribution, instead projecting the time series data to match the input dimensions of the language model~\citep{zhou2023onefitsall, cao2023tempo, chang2023llm4ts}.

In this work, we attempt to align the input distribution of time series with the word embedding of LLMs. Therefore, we propose a cross-modal match module to deal with this problem. Specifically, given a multivariate time series $I\in \mathbb{R}^{T \times C}$ as input, where $T$ is the input sequence length and $C$ is the number of variants, we first use the embedding layer similar to \citep{liu2023itransformer}, followed by Multi-head Self Attention (MHSA) to get the projected time tokens $X_{time}$:

\begin{equation}
    X_{time} = {\rm MHSA(Embedding}(I)) \in \mathbb{R}^{C \times M},
\end{equation}

\noindent 
where $M$ is the feature dimension of pre-trained LLMs. The embedding layer ${\rm Embedding}(\cdot)$ performs a channel-wise dimensional mapping from $T$ to $M$.



After that, we consider using cross-attention to align $X_{time}$ from the temporal modality and the word embedding dictionaries $\mathcal{D} \in \mathbb{R}^{|\mathcal{A}| \times M}$, where $|\mathcal{A}|$ is the size of the alphabet, to the textual modality. However, due to $|\mathcal{A}|$ is usually huge, \textit{e.g.}, 50257 in GPT2~\citep{radford2019gpt2}, directly using cross-attention incurs significant cost. Observing that semantic-similar words form ``synonym clusters'', we propose a principal word embedding extraction strategy, which uses the cluster center to represent surrounding words, to reduce the number of word entries. Specifically, we use Principal Component Analysis (PCA) to perform dimension reduction on $\mathcal{D}$ to obtain the principal word embeddings $\hat{\mathcal{D}} \in \mathbb{R}^{d \times M}$, 

\begin{equation}
    \hat{\mathcal{D}} = {\rm PCA}(\mathcal{D}),
\end{equation}

\noindent
where $d$ is a pre-defined low dimension and satisfies $ d \ll |\mathcal{A}|$.

It is worth noting that this process needs to be done only once before model training and does not incur much training overhead. We then use Multi-head Cross-Attention with $\hat{\mathcal{D}}$ as key and value, and $X_{time}$ as query to align the principal word embeddings and temporal tokens to obtain the aligned text tokens $X_{text} \in \mathbb{R}^{C \times M}$,

\begin{equation}
\begin{aligned}
       &X_{text} = {\rm Softmax}(\frac{QK^T}{\sqrt{C}})V, \\
       Q=&X_{time} W_q,K=\hat{\mathcal{D}} W_k, V=\hat{\mathcal{D}} W_v,
\end{aligned}
\end{equation}
where $W_q$, $W_k$ and $W_v \in \mathbb{R}^{M \times M}$ are the projection matrices for the query ($Q$), key ($K$), and value ($V$), respectively.

\subsection{Feature Regularization Loss}

The pre-trained weights in LLMs are based on their original textual modality data. To more effectively adapt these pre-trained weights to time series data, we align the outputs of each intermediate layer in the temporal target branch with those of the textual source branch. This alignment process, facilitated by feature regularization loss, matches the intermediate features between two branches, allowing gradients at each intermediate layer to be more effectively guided for better weight updates. Formally, given $F^l_{text}$ and $F_{time}^l$ from the outputs of the $l$-th Transformer block in the textual source branch and temporal target branches, respectively, the feature regularization loss is defined as:


\begin{equation}
    \mathcal{L}_{feature} = \sum_{l=1}^{L} \gamma^{(L-l)} {\rm sim}(\phi^{text}_l(F_{{text}}^l),  \phi^{time}_l(F_{{time}}^l)),
\end{equation}

\noindent
where $\gamma$ is a hyper-parameter that controls the loss scale from different layers, and ${\rm sim}(\cdot,\cdot)$ is a chosen similarity function, such as $L_1$ loss. Following \citep{chen2020simple}, we introduce two trainable projection layers $\phi^{text}_l(\cdot)$ and $\phi^{time}_l(\cdot)$ to transform the features from textual and temporal modalities to the shared representation space.

\subsection{Output Consistency Loss}

Building on the feature regularization loss, we further ensure consistent semantic context between the textual and temporal modalities. Output consistency loss achieves this by ensuring that the output distributions correspond effectively, resolving discrepancies in the representation space. This alignment maintains a coherent and unified semantic representation for both the time series and textual data, facilitating more accurate and reliable model predictions. Specifically, given the outputs $Y_{text}$ and $Y_{time}$ from the textual source branch and temporal target branch respectively, the output consistency loss is defined as:

\begin{equation}
    \mathcal{L}_{output} = {\rm sim}({Y}_{{text}}, Y_{time}).
\end{equation}

\noindent

\subsection{Parameter Efficient Training}
To avoid catastrophic forgetting and improve training efficiency, we employ the parameter-efficient training technique to fine-tune the pre-trained LLMs. 
Specifically, for the temporal target branch, we introduce Low-rank Adaptation (LoRA)~\citep{hu2021lora} and fine-tune the positional encoding weights. The total loss during training is the weighted summation of the supervised loss $\mathcal{L}_{sup}$, the feature regularization loss $\mathcal{L}_{feature}$, and the output consistency loss $\mathcal{L}_{output}$:

\begin{equation}
    \mathcal{L}_{total} = \mathcal{L}_{sup} + \lambda_1 \mathcal{L}_{feature} + \lambda_2 \mathcal{L}_{output},
\end{equation}

\noindent
where $\lambda_1$ and $\lambda_2$ are hyper-parameters. In the inference stage, only the output of the temporal target branch will serve as the model output.

\section{Experiments}
To demonstrate the effectiveness of the \NAME, we conduct extensive experiments on various time series forecasting tasks, including long/short-term forecasting and few/zero-shot learning. Besides, we validate the model with low complexity, highlighting its efficiency in practical applications.


\renewcommand{\arraystretch}{0.8}
\begin{table*}[htb]
\setlength{\tabcolsep}{2.5pt}
\scriptsize
\centering
\begin{threeparttable}
\begin{tabular}{c|cc|cc|cc|cc|cc|cc|cc|cc|cc|cc|cc}

\toprule
 \multicolumn{1}{c}{\scalebox{1.1}{Models}} & \multicolumn{2}{c}{\NAME} & \multicolumn{2}{c}{TimeLLM$^\dagger$} & \multicolumn{2}{c}{GPT4TS$^\dagger$} & \multicolumn{2}{c}{PatchTST} & \multicolumn{2}{c}{iTransformer} & \multicolumn{2}{c}{Crossformer} & \multicolumn{2}{c}{FEDformer} & \multicolumn{2}{c}{TimesNet} & \multicolumn{2}{c}{MICN} & \multicolumn{2}{c}{DLinear} & \multicolumn{2}{c}{TiDE} \\ 


 \cmidrule(lr){2-3} \cmidrule(lr){4-5} \cmidrule(lr){6-7} \cmidrule(lr){8-9} \cmidrule(lr){10-11} \cmidrule(lr){12-13} \cmidrule(lr){14-15} \cmidrule(lr){16-17} \cmidrule(lr){18-19} \cmidrule(lr){20-21} \cmidrule(lr){22-23}

\multicolumn{1}{c}{Metric} & MSE & MAE & MSE & MAE & MSE & MAE & MSE & MAE & MSE & MAE & MSE & MAE & MSE & MAE & MSE & MAE & MSE & MAE & MSE & MAE & MSE & MAE \\

\toprule








ETTm1 & 0.395 & \best{0.390} & 0.410 & 0.409 & \second{0.389} & 0.397 & \best{0.381} & \second{0.395} & 0.407 & 0.411 & 0.502 & 0.502 & 0.448 & 0.452 & 0.400 & 0.406 & 0.392 & 0.413 & 0.403 & 0.407 & 0.412 & 0.406 \\

\midrule








ETTm2 & \best{0.281} & \best{0.321} & 0.296 & 0.340 & \second{0.285} & 0.331 & \second{0.285} & 0.327 & 0.291 & 0.335 & 1.216 & 0.707 & 0.305 & 0.349 & 0.291 & 0.333 & 0.328 & 0.382 & 0.350 & 0.401 & 0.289 & \second{0.326} \\

\midrule








ETTh1 & \best{0.432} & \best{0.428} & 0.460 & 0.449 & 0.447 & 0.436 & 0.450 & 0.441 & 0.455 & 0.448 & 0.620 & 0.572 & \second{0.440} & 0.460 & 0.458 & 0.450 & 0.558 & 0.535 & 0.456 & 0.452 & 0.445 & \second{0.432} \\

\midrule








ETTh2 & \best{0.349} & \best{0.382} & 0.389 & 0.408 & 0.381 & 0.408 & \second{0.366} & \second{0.394} & 0.381 & 0.405 & 0.942 & 0.684 & 0.437 & 0.449 & 0.414 & 0.427 & 0.587 & 0.525 & 0.559 & 0.515 & 0.611 & 0.550 \\

\midrule








Weather & \second{0.250} & \best{0.274} & 0.274 & 0.290 & 0.264 & 0.284 & 0.258 & 0.280 & 0.257 & \second{0.279} & 0.259 & 0.315 & 0.309 & 0.360 & 0.259 & 0.287 & \best{0.242} & 0.299 & 0.265 & 0.317 & 0.271 & 0.320 \\

\midrule








ECL & \best{0.175} & \best{0.265} & 0.223 & 0.309 & 0.205 & 0.290 & 0.216 & 0.304 & \second{0.178} & \second{0.270} & 0.244 & 0.334 & 0.214 & 0.327 & 0.192 & 0.295 & 0.186 & 0.294 & 0.212 & 0.300 & 0.251 & 0.344 \\

\midrule








Traffic & \second{0.439} & \best{0.281} & 0.541 & 0.358 & 0.488 & 0.317 & 0.555 & 0.361 & \best{0.428} & \second{0.282} & 0.550 & 0.304 & 0.610 & 0.376 & 0.620 & 0.336 & 0.541 & 0.315 & 0.625 & 0.383 & 0.760 & 0.473 \\


\toprule

\end{tabular}
\begin{tablenotes}
    \footnotesize
    \item[$\dagger$] We utilize their official codebase with the same experimental setup as ours, including input length and a GPT2 model with 6 layers, to ensure the fairness of the results. Other results are obtained from \cite{liu2023itransformer}.
\end{tablenotes}
\caption{Multivariate long-term forecasting results. The input sequence length $T$ is set to 96 for all baselines. The best and second best results are in \textcolor{red}{\textbf{bold}} and \textcolor{bl}{\underline{underlined}}.} 
\label{tab::long-term}
\end{threeparttable}
\end{table*}

\renewcommand{\arraystretch}{0.8}

\begin{table*}[htb]
\setlength{\tabcolsep}{5pt}
\scriptsize
\centering
\begin{threeparttable}
\begin{tabular}{c|c|cccccccccccc}
\toprule

\multicolumn{2}{c}{\scalebox{1.1}{Models}} & \NAME & TimeLLM & GPT4TS & PatchTST & ETSformer & FEDformer & Autoformer & TimesNet & TCN & N-HiTS & N-BEATS & DLinear \\ 


\toprule

\multirow{3}{*}{\rotatebox[origin=c]{90}{Yearly}}

& \scalebox{0.8}{SMAPE} & \best{13.351} & 13.419 & 13.531 & 13.477 & 18.009 & 13.728 & 13.974 & \second{13.387} & 14.920 & 13.418 & 13.436 & 16.965 \\

& \scalebox{0.8}{MASE}  & \second{3.003} & 3.005 & 3.015  & 3.019  & 4.487  & 3.048 & 3.134 & \best{2.996}  & 3.364 & 3.045  & 3.043  & 4.283  \\

& \scalebox{0.8}{OWA}   & \best{0.786}  & 0.789 & 0.793  & \second{0.792}  & 1.115  & 0.803 & 0.822 & \best{0.786}  & 0.880 & 0.793  & 0.794  & 1.058  \\

\midrule

\multirow{3}{*}{\rotatebox[origin=c]{90}{Quarterly}}

& \scalebox{0.8}{SMAPE} & \best{9.990}  & 10.110 & 10.177 & 10.380 & 13.376 & 10.792 & 11.338 & \second{10.100} & 11.122 & 10.202 & 10.124 & 12.145 \\

& \scalebox{0.8}{MASE}  & \best{1.164}  & 1.178 & 1.194  & 1.233  & 1.906  & 1.283  & 1.365  & 1.182  & 1.360  & 1.194  & \second{1.169}  & 1.520  \\

& \scalebox{0.8}{OWA}   & \best{0.878}  & 0.889 & 0.898  & 0.921  & 1.302  & 0.958  & 1.012  & 0.890  & 1.001  & 0.899  & \second{0.886}  & 1.106  \\

\midrule

\multirow{3}{*}{\rotatebox[origin=c]{90}{Monthly}}

& \scalebox{0.8}{SMAPE} & \best{12.643} & 12.980 & 12.894 & 12.959 & 14.588 & 14.260 & 13.958 & 12.679 & 15.626 & 12.791 & \second{12.677} & 13.514 \\

& \scalebox{0.8}{MASE}  & \best{0.922}  & 0.963 & 0.956  & 0.970  & 1.368  & 1.102  & 1.103  & \second{0.933}  & 1.274  & 0.969  & 0.937  & 1.037  \\

& \scalebox{0.8}{OWA}   & \best{0.872}  & 0.903 & 0.897  & 0.905  & 1.149  & 1.012  & 1.002  & \second{0.878}  & 1.141  & 0.899  & 0.880  & 0.956  \\

\midrule

\multirow{3}{*}{\rotatebox[origin=c]{90}{Others}}

& \scalebox{0.8}{SMAPE} & \best{4.552}  & 4.795 & 4.940  & 4.952  & 7.267  & 4.954  & 5.485  & \second{4.891}  & 7.186  & 5.061  & 4.925  & 6.709  \\

& \scalebox{0.8}{MASE}  & \best{3.092}  & 3.178 & 3.228  & 3.347  & 5.240  & 3.264  & 3.865  & 3.302  & 4.677  & \second{3.216}  & 3.391  & 4.953  \\

& \scalebox{0.8}{OWA}   & \best{0.967}  & \second{1.006} & 1.029  & 1.049  & 1.591  & 1.036  & 1.187  & 1.035  & 1.494  & 1.040  & 1.053  & 1.487  \\








\bottomrule


\end{tabular}

\caption{Short-term forecasting results on M4 dataset. The input length and prediction length are set to $[12, 96]$ and $[6, 48]$, respectively.}
\label{tab::short-term}
\end{threeparttable}
\end{table*}
\renewcommand{\arraystretch}{1}


\paragraph{\textbf{Baselines.}} We carefully select representative baselines from the recent time series forecasting landscape, including the following categories: (1) LLMs-based models: TimeLLM \citep{jin2023timellm} and GPT4TS \citep{zhou2023onefitsall}; (2) Transformer-based models: PatchTST~\citep{nie2022pathtst}, iTransformer~\citep{liu2023itransformer}, Crossformer~\citep{zhang2022crossformer}, ETSformer~\citep{woo2022etsformer}, FEDformer~\citep{zhou2022fedformer} and Autoformer~\citep{wu2021autoformer}; (3) CNN-based models: TCN~\citep{BaiTCN2018}, MICN~\citep{wang2022micn} and TimesNet~\citep{wu2023timesnet}; (4) MLP-based models: DLinear~\citep{zeng2023dlinear} and TiDE~\citep{das2023tide}. Besides, N-HiTS \citep{challu2022nhits} and N-BEATS \citep{oreshkin2019nbeats} are included for short-term forecasting.

\paragraph{Implementation Details.} Following \citep{zhou2023onefitsall}, we use pre-trained GPT2 based model \citep{radford2019gpt2} with the first 6 Transformer layers as our backbone. Optimization is conducted using the Adam optimizer \citep{kingma2014adam}, with a learning rate of $0.0005$. For the total loss function, we set the hyper-parameters $\gamma=0.8$, $\lambda_1=1$ and $\lambda_2=0.01$. In terms of loss functions for long-term forecasting, we apply L1 loss across all three loss types for ETT datasets, while for the other three datasets, smooth L1 loss is utilized. For short-term forecasting, we compute supervised loss with SMAPE, modal consistency loss with MASE, and feature regularization loss with smooth L1 loss, respectively. 

\subsection{Long-term Forecasting}

\paragraph{\textbf{Setups.}} We conduct experiments on seven widely-used real-world datasets, including the Electricity Transformer Temperature (ETT) dataset with its four subsets (ETTh1, ETTh2, ETTm1, ETTm2), Weather, ECL, and Traffic \citep{wu2021autoformer}. The input time series length $T$ is fixed as $96$ for a fair comparison, and we adopt four distinct prediction horizons $H \in \{96, 192, 336, 720\}$. Consistent with prior works~\cite{li2024foundts}, the Mean Square Error (MSE) and Mean Absolute Error (MAE) are chosen as evaluation metrics. Following TBF's~\cite{qiu2024tfb} setting, we do not use the drop last trick to ensure fair comparison.

\paragraph{\textbf{Results.}} Comprehensive long-term forecasting results are presented in \cref{tab::long-term}, where all the results are averaged from different prediction lengths. Notably, our approach reduces MSE/MAE by 7.05\%/6.53\% compared to the state-of-the-art Transformer-based model PatchTST. In comparison with the LLM-powered method TimeLLM, we observe a reduction of 5.98\%/5.34\% in MSE/MAE. Moreover, our improvements are substantial against other baseline methods, exceeding 10\% in most cases. 


\subsection{Short-term Forecasting}

\paragraph{\textbf{Setups.}} We adopt the M4 datasets \citep{makridakis2018m4}, which comprise univariate marketing data collected yearly, quarterly, and monthly. In this case, the prediction horizons are comparatively short, ranging in $[6, 48]$. Correspondingly, the input lengths are set to be twice the size of the prediction horizons. The evaluation metrics are symmetric mean absolute percentage error (SMAPE), mean absolute scaled error (MSAE), and overall weighted average (OWA).

\paragraph{\textbf{Results.}} As shown in \cref{tab::short-term}, our method demonstrates superior performance in short-term forecasting across various evaluation metrics. In comparison with TimesNet, currently the leading method in short-term forecasting, our model achieves a 1\% overall improvement in performance.


\subsection{Few/zero-shot Learning}

LLMs have demonstrated remarkable performance in both few-shot and zero-shot tasks. The capabilities of few-shot and zero-shot learning are critically important for general time series forecasting models \citep{brown2020llm_few, liu2023few_shot, kojima2022zero_shot}. To thoroughly assess the generalized ability of our method in time series forecasting, we conduct experiments under few-shot and zero-shot learning settings. In few-shot learning, only a small ratio of the training data is utilized. For zero-shot learning, the model trained on one dataset is directly employed for testing on another dataset without any additional training. All the results are averaged from 4 different prediction lengths $H \in \{96, 192, 336, 720\}$. 

\renewcommand{\arraystretch}{0.7}
\begin{table}[!t]
    \setlength{\tabcolsep}{4pt}
    \centering
    \scriptsize
    \begin{threeparttable}
    \begin{tabular}{c|cc|cc|cc|cc}
    \toprule
         \multicolumn{1}{c}{\multirow{2}{*}{\scalebox{1.2}{Models}}} & \multicolumn{2}{c}{ETTm1} & \multicolumn{2}{c}{ETTm2} & \multicolumn{2}{c}{ETTh1} & \multicolumn{2}{c}{ETTh2} \\

          \cmidrule(lr){2-3} \cmidrule(lr){4-5} \cmidrule(lr){6-7} \cmidrule(lr){8-9}
         
         \multicolumn{1}{c}{} & MSE & MAE & MSE & MAE & MSE & MAE & MSE & MAE \\
    \toprule

    TiDE & \second{0.515} & \second{0.469} & 0.303 & 0.337 & 0.779 & 0.604 & \second{0.421} & \second{0.428} \\

    DLinear & 0.567 & 0.499 & 0.329 & 0.382 & \second{0.647} & 0.552 & 0.441 & 0.458 \\

    MICN & 0.970 & 0.674 & 1.073 & 0.716 & 1.405 & 0.814 & 2.533 & 1.158 \\

    TimesNet & 0.673 & 0.534 & 0.321 & 0.354 & 0.865 & 0.625 & 0.476 & 0.463 \\

    FEDformer & 0.696 & 0.572 & 0.356 & 0.392 & 0.750 & 0.607 & 0.553 & 0.525 \\

    Crossformer & 1.340 & 0.848 & 1.985 & 1.048 & 1.744 & 0.914 & 3.139 & 1.378 \\

    PatchTST & 0.557 & 0.483 & \best{0.295} & \second{0.334} & 0.683 & \second{0.546} & 0.550 & 0.487 \\

    GPT4TS & 0.608 & 0.500 & 0.303 & 0.336 & 0.689 & 0.555 & 0.579 & 0.497 \\

    TimeLLM   & 0.636 & 0.512 & 0.348 & 0.343 & 0.765 & 0.584 & 0.589 & 0.498 \\

    \NAME & \best{0.504} & \best{0.462} & \second{0.302} & \best{0.330} & \best{0.644} & \best{0.541} & \best{0.419} & \best{0.427} \\

    \bottomrule

    \end{tabular}
    \caption{Few-shot learning results on 10\% training data of ETT datasets.}
    \label{tab:few-shot}
    \end{threeparttable}
\end{table}
\renewcommand{\arraystretch}{1}

\paragraph{\textbf{Few-shot Learning.}} We conduct few-shot experiments on four ETT datasets. Specifically, for each dataset, we utilize only the first 10\% of the training data. This constrained data scenario presents a considerable challenge, testing the ability of the model to learn effectively with limited information. \cref{tab:few-shot} demonstrates that our method outperforms other baselines, highlighting its robustness in the few-shot setting. Compared with TimeLLM and PatchTST, our method achieves an average reduction of 8\% and 9\%, respectively.

\renewcommand{\arraystretch}{0.7}
\begin{table}[!t]
\setlength{\tabcolsep}{4pt}
\scriptsize
    \centering
    \begin{threeparttable}
    \begin{tabular}{c|cc|cc|cc|cc}
    
    \toprule



    \multicolumn{1}{c}{\multirow{2}{*}{\scalebox{1.2}{Models}}} & \multicolumn{2}{c}{h1 $\rightarrow$ m1} & \multicolumn{2}{c}{h1 $\rightarrow$ m2} & \multicolumn{2}{c}{h2 $\rightarrow$ m1} & \multicolumn{2}{c}{h2 $\rightarrow$ m2} \\
    
    \cmidrule(lr){2-3} \cmidrule(lr){4-5} \cmidrule(lr){6-7} \cmidrule(lr){8-9}

    \multicolumn{1}{c}{} & MSE & MAE & MSE & MAE & MSE & MAE & MSE & MAE \\

    \toprule

    TiDE & 0.774 & \best{0.574} & \best{0.314} & \best{0.355} & 0.841 & 0.590 & \second{0.321} & \second{0.364} \\

    DLinear & \second{0.760} & 0.577 & 0.399 & 0.439 & 0.778 & 0.594 & 0.496 & 0.496 \\

    MICN & 1.439 & 0.780 & 2.428 & 1.236 & \second{0.764} & 0.601 & 0.527 & 0.519  \\

    TimesNet & 0.794 & \second{0.575} & 0.339 & 0.370 & 1.286 & 0.705 & 0.361 & 0.390 \\

    FEDformer & 0.765 & 0.588 & 0.357 & 0.403 & \best{0.741} & \second{0.588} & 0.365 & 0.405 \\

    Crossformer & 0.999 & 0.736 & 1.120 & 0.789 & 1.195 & 0.711 & 2.043 & 1.124 \\

    PatchTST & 0.894 & 0.610 & 0.318 & 0.362 & 0.871 & 0.596 & 0.420 & 0.433 \\

    GPT4TS  & 0.798 & \best{0.574} & 0.317 & \second{0.359} & 0.920 & 0.610 & 0.331 & 0.371 \\

    TimeLLM & 0.847 & 0.565 & 0.315 & 0.357 & 0.868 & 0.595 & 0.322 & 0.363 \\

    \NAME & \best{0.755} & \best{0.574} & \second{0.316} & \best{0.355} & 0.836 & \best{0.586} & \best{0.319} & \best{0.360} \\

    \toprule
    
    \end{tabular}
    \caption{Zero-shot learning results on ETT datasets. ``$\blacklozenge \to \bigstar$'' indicates that models trained on the dataset $\blacklozenge$ are evaluated on a distinct dataset $\bigstar$.}
    \label{tab:zero-shot}
    \end{threeparttable}
\end{table}
\renewcommand{\arraystretch}{1}

\paragraph{\textbf{Zero-shot Learning.}} Going beyond few-shot scenarios, we further delve into zero-shot learning, where LLMs demonstrate their prowess as adept and intuitive reasoners. In this setting, models trained on one dataset $\blacklozenge$ are evaluated on an entirely different dataset $\bigstar$, without any further training. As shown in \cref{tab:zero-shot}, our method stands out for its exceptional performance, surpassing TimeLLM and PatchTST by 4\% and 9\% respectively. This indicates that our approach significantly enhances the model's capability for effective learning transfer across different domains.

\renewcommand{\arraystretch}{0.7}
\begin{table}[!t]
    \setlength{\tabcolsep}{2pt}
    \centering
    \scriptsize
    \begin{threeparttable}
    \begin{tabular}{c|cc|cc|cc}
    \toprule
    Dataset & \multicolumn{2}{c|}{GPT4TS} & \multicolumn{2}{c|}{Time-LLM} & \multicolumn{2}{c}{\NAME} \\
    \midrule
    & Time (s) & MSE / MAE & Time (s) & MSE / MAE & Time (s) & MSE / MAE \\
    \midrule
    ETTm1   & 626  & 0.329 / 0.364 & 1476 & 0.359 / 0.381 & \best{135}  & \best{0.323 / 0.349} \\
    ECL     & 8274 & 0.185 / 0.272 & 33209 & 0.204 / 0.293 & \best{251}  & \best{0.145 / 0.238} \\
    Traffic & 15067 & 0.468 / 0.307 & 62412 & 0.536 / 0.359 & \best{614}  & \best{0.407 / 0.268} \\
    Weather & 596  & 0.182 / 0.223 & 1262 & 0.195 / 0.233 & \best{123}  & \best{0.164 / 0.204} \\
    \toprule
    \end{tabular}
    \caption{Comparison of different LLM-based time series forecasting methods in terms of computation time and performance (MSE/MAE) across various datasets.}
    \label{tab:time_and_performance}
    \end{threeparttable}
\end{table}
\renewcommand{\arraystretch}{1}

\subsection{Efficiency Analysis}
We conduct experiments on four datasets: ETTm1, ECL, Traffic, and Weather. The input and prediction lengths are both set to 96. As shown in \cref{tab:time_and_performance}, our proposed \NAME shows significant improvements in both efficiency and accuracy compared with other LLM-based methods, largely due to treating each channel sequence as a token, employing an efficient fine-tuning strategy, and requiring only a single time branch during inference. 

%


\begin{figure*}[!t]
\centering
\includegraphics[width=0.95\textwidth]{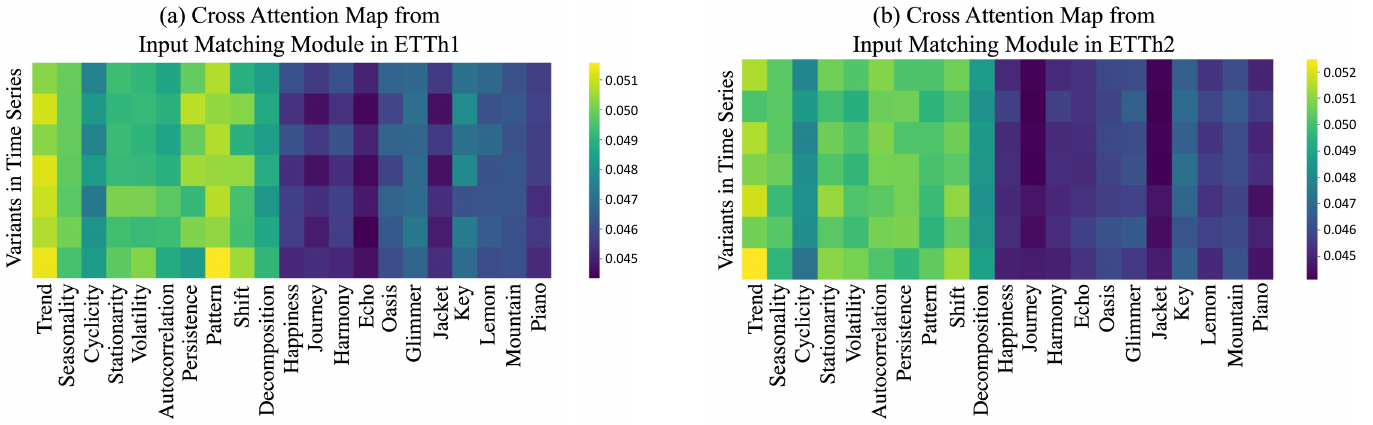}
\caption{Cross-attention maps from the Cross-Modal Match Module for ETTh1 (left) and ETTh2 (right). Each row represents a time series instance, while columns correspond to selected words, including both time-related terms (\textit{e.g.}, trend, seasonality) and general terms (\textit{e.g.}, echo, key). Each cell indicates the relevance of the respective channel to the selected word.}
\label{fig:cross-attn-map}
\end{figure*}

\section{Ablation Study}

\paragraph{Ablation on Different Loss Functions.} The feature regularization loss $\mathcal{L}_{feature}$ aligns the intermediate features between the textual source branch and the temporal target branch, while the output consistency loss $\mathcal{L}_{output}$ ensures output coherence across modalities. The supervised loss $\mathcal{L}_{sup}$ directly guides learning with ground truth data. We analyze the specific effects of each proposed loss function as detailed in \cref{tab:loss_abalation}. Employing only the supervised loss resulted in MSE/MAE of 0.446/0.438 on ETTh1 and 0.263/0.286 on Weather, respectively. The addition of feature regularization loss $\mathcal{L}_{feature}$ or output consistency loss $\mathcal{L}_{output}$ led to incremental improvements, with the best performance observed when all three losses were combined, achieving the lowest MSE and MAE on both datasets.

\begin{figure}[!t]
    \centering
    \includegraphics[width=0.96\linewidth]{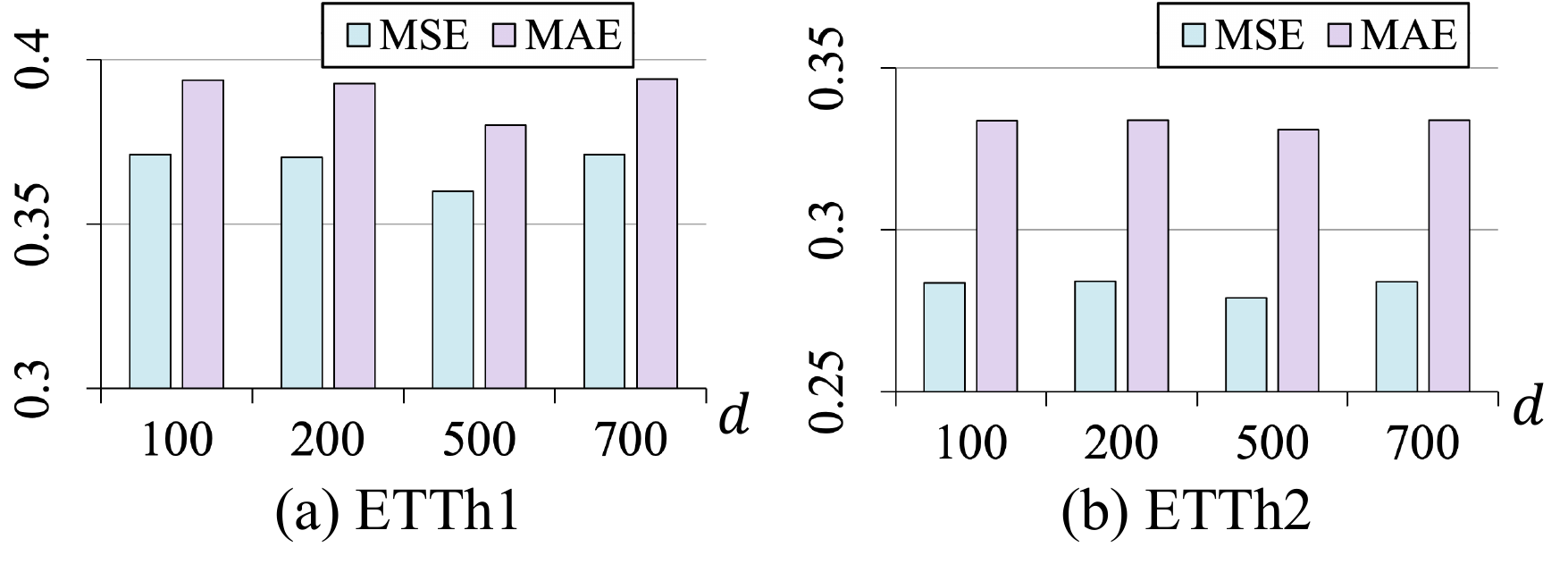}
    \caption{Ablation on different low dimension $d$ of PCA on
(a) ETTh1 and (b) ETTh2 datasets.}
    \label{fig:ablation-pca}
\end{figure}

\paragraph{Ablation on the Number of Principal Components.} 
We employ PCA to conduct dimensional reduction on the original word embeddings for efficient training. Despite the reduced cost, however, PCA may inevitably lead to information loss. In this section, we ablate the number of principal components $d$ to present the effects. The experimental results are given in ~\cref{fig:ablation-pca}. It can be seen that the performance is not that sensitive to different numbers of principal components. In addition, a smaller $d$ causes performance degradation due to the missing key information, while a larger $d$ causes information redundancy which causes learning difficulty. In practice, we chose $d = 500$, which can attain an explainable variance ratio of 88\% while achieving satisfactory performance.


\renewcommand{\arraystretch}{0.5}
\begin{table}[!t]
    \setlength{\tabcolsep}{5pt}
    \scriptsize 
    \centering
    \begin{threeparttable}
    \begin{tabular}{ccccccc} 
    \toprule
    \multirow{2}{*}{\scalebox{1.3}{$\mathcal{L}_{feature}$}} & 
    \multirow{2}{*}{\scalebox{1.3}{$\mathcal{L}_{output}$}} & 
    \multirow{2}{*}{\scalebox{1.3}{$\mathcal{L}_{sup}$}} & 
    \multicolumn{2}{c}{ETTh1} & 
    \multicolumn{2}{c}{Weather} \\
    \cmidrule(lr){4-5} \cmidrule(lr){6-7}
    &  &  & MSE & MAE & MSE & MAE \\
    \midrule
    $-$ & $-$ & \checkmark & 0.446 & 0.438 & 0.263 & 0.286 \\
    \checkmark & $-$ & \checkmark & 0.434 & 0.431 & 0.254 & 0.276 \\
    $-$ & \checkmark & \checkmark & 0.438 & \best{0.426} & 0.258 & 0.283 \\
    \checkmark & \checkmark & \checkmark & \best{0.432} & 0.428 & \best{0.250} & \best{0.274} \\
    \bottomrule
    \end{tabular}
    \caption{Ablation on different loss functions on ETTh1 and Weather datasets.}
    \label{tab:loss_abalation}
    \end{threeparttable}
    \label{tab:my_label}
\end{table}
\renewcommand{\arraystretch}{1}

\section{Discussion}
\label{sec::discussion}


\paragraph{Difference from Other Work.}
Existing LLM-based methods, such as TimeLLM~\cite{jin2023timellm} and TEST~\cite{sun2024test}, also consider the alignment between the time-modal input and the text-modal parameters by employing the cross-attention layer or contrastive learning at the input side. However, we point out that this simple scheme is not sufficiently powerful thus leading to partial alignment (\cref{fig:intro}) as well as weak generalization performance (\cref{tab:few-shot}, \cref{tab:zero-shot}). By contrast, we identify the alignment problem and further propose a multi-level cross-modal fine-tuning framework to achieve finer-grained alignment. Moreover, although the previous approaches adopt one cross-attention layer as an intuitive solution, it can raise significant computational costs due to the huge alphabet size of LLM. In this work, we introduce the offline PCA to generate synonym clusters from the huge alphabet as an alternative, allowing high efficiency while achieving better performance.


\paragraph{Interpretability on Implicit Input Alignment.}
To narrow the temporal-textual modality gap, we perform cross-attention on word embedding weights to generate aligned text tokens instead of intuitive natural language. As shown in ~\cref{fig:cross-attn-map}, we visualize the cross-attention maps from the Cross-Modal Match Module for the ETTh1 and ETTh2 datasets. Each row in the maps represents a time series instance, while columns correspond to selected words, including both time-related terms (\textit{e.g.}, trend, seasonality) and general terms (\textit{e.g.}, echo, key). Each cell indicates the relevance of the respective channel to the selected word. Our analysis reveals that the Cross-Modal Match Module effectively aligns time series tokens with word embeddings that describe temporal characteristics. The attention distributions show that time series data align well with relevant textual descriptions, indicating that our module successfully bridges the gap between temporal and textual modalities. 




\section{Conclusion}
In this work, we propose \NAME, a novel cross-modal fine-tuning framework that leverages the robust capabilities of Large Language Models (LLMs) for time series forecasting. \NAME effectively bridges the distribution discrepancy between temporal data and the textual nature of LLMs through the Cross-Modal Match Module, Feature Regularization Loss, and Output Consistency Loss. Extensive experiments across several real-world datasets validate that \NAME sets a new benchmark in both long- and short-term forecasting, demonstrating strong generalization and low computational complexity. 

\clearpage

\section{Acknowledgements}
This work is supported in part by the National Natural Science Foundation of China, under Grant (62302309, 62171248), Shenzhen Science and Technology Program (JCYJ20220818101014030, JCYJ20220818101012025), and the PCNL KEY project (PCL2023AS6-1).

\bibliography{aaai25}

\begin{thebibliography}{43}
\providecommand{\natexlab}[1]{#1}

\bibitem[{Angryk et~al.(2020)Angryk, Martens, Aydin, Kempton, Mahajan, Basodi, Ahmadzadeh, Cai, Filali~Boubrahimi, Hamdi et~al.}]{angryk2020weather}
Angryk, R.~A.; Martens, P.~C.; Aydin, B.; Kempton, D.; Mahajan, S.~S.; Basodi, S.; Ahmadzadeh, A.; Cai, X.; Filali~Boubrahimi, S.; Hamdi, S.~M.; et~al. 2020.
\newblock Multivariate time series dataset for space weather data analytics.
\newblock \emph{Scientific data}, 7(1): 227.

\bibitem[{Bai, Kolter, and Koltun(2018)}]{BaiTCN2018}
Bai, S.; Kolter, J.~Z.; and Koltun, V. 2018.
\newblock An empirical evaluation of generic convolutional and recurrent networks for sequence modeling.
\newblock \emph{arXiv preprint arXiv:1803.01271}.

\bibitem[{Brown et~al.(2020)Brown, Mann, Ryder, Subbiah, Kaplan, Dhariwal, Neelakantan, Shyam, Sastry, Askell et~al.}]{brown2020llm_few}
Brown, T.; Mann, B.; Ryder, N.; Subbiah, M.; Kaplan, J.~D.; Dhariwal, P.; Neelakantan, A.; Shyam, P.; Sastry, G.; Askell, A.; et~al. 2020.
\newblock Language models are few-shot learners.
\newblock \emph{Advances in Neural Information Processing Systems}, 33: 1877--1901.

\bibitem[{Cao et~al.(2024)Cao, Jia, Arik, Pfister, Zheng, Ye, and Liu}]{cao2023tempo}
Cao, D.; Jia, F.; Arik, S.~O.; Pfister, T.; Zheng, Y.; Ye, W.; and Liu, Y. 2024.
\newblock T{EMPO}: Prompt-based Generative Pre-trained Transformer for Time Series Forecasting.
\newblock \emph{International Conference on Learning Representations}.

\bibitem[{Challu et~al.(2022)Challu, Olivares, Oreshkin, Garza, Mergenthaler, and Dubrawski}]{challu2022nhits}
Challu, C.; Olivares, K.~G.; Oreshkin, B.~N.; Garza, F.; Mergenthaler, M.; and Dubrawski, A. 2022.
\newblock N-{H}i{T}S: Neural Hierarchical Interpolation for Time Series Forecasting.
\newblock \emph{arXiv preprint arXiv:2201.12886}.

\bibitem[{Chang, Peng, and Chen(2023)}]{chang2023llm4ts}
Chang, C.; Peng, W.-C.; and Chen, T.-F. 2023.
\newblock L{LM}4{TS}: Two-stage fine-tuning for time-series forecasting with pre-trained llms.
\newblock \emph{arXiv preprint arXiv:2308.08469}.

\bibitem[{Chen et~al.(2020)Chen, Kornblith, Norouzi, and Hinton}]{chen2020simple}
Chen, T.; Kornblith, S.; Norouzi, M.; and Hinton, G. 2020.
\newblock A simple framework for contrastive learning of visual representations.
\newblock In \emph{International Conference on Machine Learning}, 1597--1607. PMLR.

\bibitem[{Dai et~al.(2024)Dai, Wu, Liu, Li, Bao, Jiang, and Xia}]{dai2023periodicity}
Dai, T.; Wu, B.; Liu, P.; Li, N.; Bao, J.; Jiang, Y.; and Xia, S.-T. 2024.
\newblock Periodicity Decoupling Framework for Long-term Series Forecasting.
\newblock In \emph{International Conference on Learning Representations}.

\bibitem[{Das et~al.(2023)Das, Kong, Leach, Sen, and Yu}]{das2023tide}
Das, A.; Kong, W.; Leach, A.; Sen, R.; and Yu, R. 2023.
\newblock Long-term Forecasting with {T}i{DE}: Time-series Dense Encoder.
\newblock \emph{arXiv preprint arXiv:2304.08424}.

\bibitem[{Demirel et~al.(2012)Demirel, Zaim, {\c{C}}ali{\c{s}}kan, and {\"O}zuyar}]{demirel2012forecasting}
Demirel, {\"O}.~F.; Zaim, S.; {\c{C}}ali{\c{s}}kan, A.; and {\"O}zuyar, P. 2012.
\newblock Forecasting natural gas consumption in Istanbul using neural networks and multivariate time series methods.
\newblock \emph{Turkish Journal of Electrical Engineering and Computer Sciences}, 20(5): 695--711.

\bibitem[{Hassid et~al.(2024)Hassid, Remez, Nguyen, Gat, Conneau, Kreuk, Copet, Defossez, Synnaeve, Dupoux et~al.}]{hassid2024speak}
Hassid, M.; Remez, T.; Nguyen, T.~A.; Gat, I.; Conneau, A.; Kreuk, F.; Copet, J.; Defossez, A.; Synnaeve, G.; Dupoux, E.; et~al. 2024.
\newblock Textually pretrained speech language models.
\newblock \emph{Advances in Neural Information Processing Systems}, 36.

\bibitem[{Hu et~al.(2021)Hu, Shen, Wallis, Allen-Zhu, Li, Wang, Wang, and Chen}]{hu2021lora}
Hu, E.~J.; Shen, Y.; Wallis, P.; Allen-Zhu, Z.; Li, Y.; Wang, S.; Wang, L.; and Chen, W. 2021.
\newblock Lo{RA}: Low-rank adaptation of large language models.
\newblock \emph{arXiv preprint arXiv:2106.09685}.

\bibitem[{Jin et~al.(2024)Jin, Wang, Ma, Chu, Zhang, Shi, Chen, Liang, Li, Pan et~al.}]{jin2023timellm}
Jin, M.; Wang, S.; Ma, L.; Chu, Z.; Zhang, J.~Y.; Shi, X.; Chen, P.-Y.; Liang, Y.; Li, Y.-F.; Pan, S.; et~al. 2024.
\newblock Time-{LLM}: Time series forecasting by reprogramming large language models.
\newblock \emph{International Conference on Learning Representations}.

\bibitem[{Jin et~al.(2023)Jin, Hu, Chen, Miao, Hu, and Zhao}]{jin2023cross}
Jin, Y.; Hu, G.; Chen, H.; Miao, D.; Hu, L.; and Zhao, C. 2023.
\newblock Cross-modal distillation for speaker recognition.
\newblock In \emph{Proceedings of the AAAI Conference on Artificial Intelligence}, volume~37, 12977--12985.

\bibitem[{Kiela et~al.(2019)Kiela, Bhooshan, Firooz, Perez, and Testuggine}]{kiela2019vision}
Kiela, D.; Bhooshan, S.; Firooz, H.; Perez, E.; and Testuggine, D. 2019.
\newblock Supervised multimodal bitransformers for classifying images and text.
\newblock \emph{arXiv preprint arXiv:1909.02950}.

\bibitem[{Kingma and Ba(2014)}]{kingma2014adam}
Kingma, D.~P.; and Ba, J. 2014.
\newblock Adam: A method for stochastic optimization.
\newblock \emph{arXiv preprint arXiv:1412.6980}.

\bibitem[{Kojima et~al.(2022)Kojima, Gu, Reid, Matsuo, and Iwasawa}]{kojima2022zero_shot}
Kojima, T.; Gu, S.~S.; Reid, M.; Matsuo, Y.; and Iwasawa, Y. 2022.
\newblock Large language models are zero-shot reasoners.
\newblock \emph{Advances in Neural Information Processing Systems}, 35: 22199--22213.

\bibitem[{Li et~al.(2024)Li, Qiu, Chen, Wang, Cheng, Shu, Hu, Guo, Zhou, Wen et~al.}]{li2024foundts}
Li, Z.; Qiu, X.; Chen, P.; Wang, Y.; Cheng, H.; Shu, Y.; Hu, J.; Guo, C.; Zhou, A.; Wen, Q.; et~al. 2024.
\newblock FoundTS: Comprehensive and Unified Benchmarking of Foundation Models for Time Series Forecasting.
\newblock \emph{arXiv preprint arXiv:2410.11802}.

\bibitem[{Liu et~al.(2023{\natexlab{a}})Liu, Wu, Li, Dai, Lei, Bao, Jiang, and Xia}]{liu2023wftnet}
Liu, P.; Wu, B.; Li, N.; Dai, T.; Lei, F.; Bao, J.; Jiang, Y.; and Xia, S.-T. 2023{\natexlab{a}}.
\newblock W{FTN}et: Exploiting Global and Local Periodicity in Long-term Time Series Forecasting.
\newblock \emph{IEEE International Conference on Acoustics, Speech and Signal Processing}.

\bibitem[{Liu et~al.(2023{\natexlab{b}})Liu, McDuff, Kovacs, Galatzer-Levy, Sunshine, Zhan, Poh, Liao, Di~Achille, and Patel}]{liu2023few_shot}
Liu, X.; McDuff, D.; Kovacs, G.; Galatzer-Levy, I.; Sunshine, J.; Zhan, J.; Poh, M.-Z.; Liao, S.; Di~Achille, P.; and Patel, S. 2023{\natexlab{b}}.
\newblock Large Language Models are Few-Shot Health Learners.
\newblock \emph{arXiv preprint arXiv:2305.15525}.

\bibitem[{Liu et~al.(2024)Liu, Hu, Zhang, Wu, Wang, Ma, and Long}]{liu2023itransformer}
Liu, Y.; Hu, T.; Zhang, H.; Wu, H.; Wang, S.; Ma, L.; and Long, M. 2024.
\newblock i{T}ransformer: Inverted transformers are effective for time series forecasting.
\newblock \emph{International Conference on Learning Representations}.

\bibitem[{Makridakis, Spiliotis, and Assimakopoulos(2018)}]{makridakis2018m4}
Makridakis, S.; Spiliotis, E.; and Assimakopoulos, V. 2018.
\newblock The M4 Competition: Results, findings, conclusion and way forward.
\newblock \emph{International Journal of Forecasting}, 34(4): 802--808.

\bibitem[{Mikolov et~al.(2013)Mikolov, Chen, Corrado, and Dean}]{mikolov2013efficient}
Mikolov, T.; Chen, K.; Corrado, G.; and Dean, J. 2013.
\newblock Efficient estimation of word representations in vector space.
\newblock \emph{arXiv preprint arXiv:1301.3781}.

\bibitem[{Nie et~al.(2023)Nie, H.~Nguyen, Sinthong, and Kalagnanam}]{nie2022pathtst}
Nie, Y.; H.~Nguyen, N.; Sinthong, P.; and Kalagnanam, J. 2023.
\newblock A Time Series is Worth 64 Words: Long-term Forecasting with Transformers.
\newblock In \emph{International Conference on Learning Representations}.

\bibitem[{Oreshkin et~al.(2019)Oreshkin, Carpov, Chapados, and Bengio}]{oreshkin2019nbeats}
Oreshkin, B.~N.; Carpov, D.; Chapados, N.; and Bengio, Y. 2019.
\newblock N-{BEATS}: Neural basis expansion analysis for interpretable time series forecasting.
\newblock \emph{International Conference on Learning Representations}.

\bibitem[{Patton(2013)}]{patton2013copula}
Patton, A. 2013.
\newblock Copula methods for forecasting multivariate time series.
\newblock \emph{Handbook of economic forecasting}, 2: 899--960.

\bibitem[{Qiu et~al.(2024{\natexlab{a}})Qiu, Hu, Zhou, Wu, Du, Zhang, Guo, Zhou, Jensen, Sheng et~al.}]{qiu2024tfb}
Qiu, X.; Hu, J.; Zhou, L.; Wu, X.; Du, J.; Zhang, B.; Guo, C.; Zhou, A.; Jensen, C.~S.; Sheng, Z.; et~al. 2024{\natexlab{a}}.
\newblock Tfb: Towards comprehensive and fair benchmarking of time series forecasting methods.
\newblock \emph{arXiv preprint arXiv:2403.20150}.

\bibitem[{Qiu et~al.(2024{\natexlab{b}})Qiu, Wu, Lin, Guo, Hu, and Yang}]{duet}
Qiu, X.; Wu, X.; Lin, Y.; Guo, C.; Hu, J.; and Yang, B. 2024{\natexlab{b}}.
\newblock DUET: Dual Clustering Enhanced Multivariate Time Series Forecasting.
\newblock \emph{arXiv preprint arXiv:2412.10859}.

\bibitem[{Radford et~al.(2019)Radford, Wu, Child, Luan, Amodei, Sutskever et~al.}]{radford2019gpt2}
Radford, A.; Wu, J.; Child, R.; Luan, D.; Amodei, D.; Sutskever, I.; et~al. 2019.
\newblock Language models are unsupervised multitask learners.

\bibitem[{Shen et~al.(2023)Shen, Li, Dery, Staten, Khodak, Neubig, and Talwalkar}]{shen2023orca}
Shen, J.; Li, L.; Dery, L.~M.; Staten, C.; Khodak, M.; Neubig, G.; and Talwalkar, A. 2023.
\newblock Cross-modal fine-tuning: Align then refine.
\newblock In \emph{International Conference on Machine Learning}, 31030--31056. PMLR.

\bibitem[{Sun et~al.(2024)Sun, Li, Li, and Hong}]{sun2024test}
Sun, C.; Li, H.; Li, Y.; and Hong, S. 2024.
\newblock {TEST}: Text Prototype Aligned Embedding to Activate {LLM}'s Ability for Time Series.
\newblock In \emph{The International Conference on Learning Representations}.

\bibitem[{Verma et~al.(2024)Verma, Choi, Sharma, Watson-Daniels, Oh, and Kumar}]{verma2024vision1}
Verma, G.; Choi, M.; Sharma, K.; Watson-Daniels, J.; Oh, S.; and Kumar, S. 2024.
\newblock Mysterious Projections: Multimodal LLMs Gain Domain-Specific Visual Capabilities Without Richer Cross-Modal Projections.
\newblock \emph{arXiv preprint arXiv:2402.16832}.

\bibitem[{Vinod, Chen, and Das(2023)}]{vinod2023bio}
Vinod, R.; Chen, P.-Y.; and Das, P. 2023.
\newblock Reprogramming pretrained language models for protein sequence representation learning.
\newblock \emph{arXiv preprint arXiv:2301.02120}.

\bibitem[{Wang et~al.(2022)Wang, Peng, Huang, Wang, Chen, and Xiao}]{wang2022micn}
Wang, H.; Peng, J.; Huang, F.; Wang, J.; Chen, J.; and Xiao, Y. 2022.
\newblock M{ICN}: Multi-scale local and global context modeling for long-term series forecasting.
\newblock In \emph{International Conference on Learning Representations}.

\bibitem[{Wen et~al.(2022)Wen, Zhou, Zhang, Chen, Ma, Yan, and Sun}]{wen2022survey}
Wen, Q.; Zhou, T.; Zhang, C.; Chen, W.; Ma, Z.; Yan, J.; and Sun, L. 2022.
\newblock Transformers in time series: A survey.
\newblock \emph{arXiv preprint arXiv:2202.07125}.

\bibitem[{Woo et~al.(2022)Woo, Liu, Sahoo, Kumar, and Hoi}]{woo2022etsformer}
Woo, G.; Liu, C.; Sahoo, D.; Kumar, A.; and Hoi, S. C.~H. 2022.
\newblock E{TS}former: Exponential Smoothing Transformers for Time-series Forecasting.
\newblock \emph{arXiv preprint arXiv:2202.01381}.

\bibitem[{Wu et~al.(2023)Wu, Hu, Liu, Zhou, Wang, and Long}]{wu2023timesnet}
Wu, H.; Hu, T.; Liu, Y.; Zhou, H.; Wang, J.; and Long, M. 2023.
\newblock Times{N}et: Temporal 2D-Variation Modeling for General Time Series Analysis.
\newblock In \emph{International Conference on Learning Representations}.

\bibitem[{Wu et~al.(2021)Wu, Xu, Wang, and Long}]{wu2021autoformer}
Wu, H.; Xu, J.; Wang, J.; and Long, M. 2021.
\newblock Autoformer: Decomposition Transformers with {Auto-Correlation} for Long-Term Series Forecasting.
\newblock In \emph{Advances in Neural Information Processing Systems}.

\bibitem[{Xiao et~al.(2021)Xiao, Qiu, Li, Hsieh, and Tang}]{xiao2021bio1}
Xiao, Y.; Qiu, J.; Li, Z.; Hsieh, C.-Y.; and Tang, J. 2021.
\newblock Modeling protein using large-scale pretrain language model.
\newblock \emph{arXiv preprint arXiv:2108.07435}.

\bibitem[{Zeng et~al.(2023)Zeng, Chen, Zhang, and Xu}]{zeng2023dlinear}
Zeng, A.; Chen, M.; Zhang, L.; and Xu, Q. 2023.
\newblock Are transformers effective for time series forecasting?
\newblock In \emph{Proceedings of the AAAI conference on artificial intelligence}, volume~37, 11121--11128.

\bibitem[{Zhang and Yan(2023)}]{zhang2022crossformer}
Zhang, Y.; and Yan, J. 2023.
\newblock Crossformer: Transformer utilizing cross-dimension dependency for multivariate time series forecasting.
\newblock In \emph{International Conference on Learning Representations}.

\bibitem[{Zhou et~al.(2022)Zhou, Ma, Wen, Wang, Sun, and Jin}]{zhou2022fedformer}
Zhou, T.; Ma, Z.; Wen, Q.; Wang, X.; Sun, L.; and Jin, R. 2022.
\newblock F{ED}former: Frequency enhanced decomposed transformer for long-term series forecasting.
\newblock In \emph{International Conference on Machine Learning}, 27268--27286. PMLR.

\bibitem[{Zhou et~al.(2023)Zhou, Niu, Wang, Sun, and Jin}]{zhou2023onefitsall}
Zhou, T.; Niu, P.; Wang, X.; Sun, L.; and Jin, R. 2023.
\newblock {One Fits All}: Power General Time Series Analysis by Pretrained LM.
\newblock \emph{Advances in Neural Information Processing Systems}, 36.

\end{thebibliography}

\end{document}